\documentclass[conference]{IEEEtran}
\IEEEoverridecommandlockouts
\usepackage{cite}
\usepackage{amsmath,amssymb,amsfonts}
\usepackage{algorithmic}
\usepackage{graphicx}
\usepackage{textcomp}
\usepackage{xcolor}
\usepackage[capitalise]{cleveref}
\def\BibTeX{{\rm B\kern-.05em{\sc i\kern-.025em b}\kern-.08em
    T\kern-.1667em\lower.7ex\hbox{E}\kern-.125emX}}
\usepackage{listings}
\usepackage{flushend}

\begin{document}

\title{OV-HHIR: Open Vocabulary Human Interaction Recognition Using Cross-modal Integration of Large Language Models
\thanks{The research reported in this paper was supported by the BMBF in the project VidGenSense (01IW21003) and by Carl-Zeiss Stiftung under the Sustainable Embedded AI project (P2021-02-009).

}
}

%\author{\IEEEauthorblockN{1\textsuperscript{st} Lala Shakti Swarup Ray}
%\IEEEauthorblockA{\textit{DFKI GmbH} \\
%Kaiserslautern, Germany \\
%lala\_shakti\_swarup.ray@dfk.de}
%\and
%\IEEEauthorblockN{2\textsuperscript{nd} Bo Zhou}
%\IEEEauthorblockA{\textit{DFKI GmbH,} \\
%\textit{RPTU Kaiserslautern-Landau} \\
%Kaiserslautern, Germany \\
%bo.zhou@dfki.de}
%\and
%\IEEEauthorblockN{3\textsuperscript{rd} Sungho Suh}
%\IEEEauthorblockA{\textit{DFKI GmbH} \\
%\textit{RPTU Kaiserslautern-Landau} \\
%Kaiserslautern, Germany \\
%sunghho.suh@dfki.de}
%\and
%\IEEEauthorblockN{4\textsuperscript{th} Paul Lukowicz}
%\IEEEauthorblockA{\textit{DFKI GmbH} \\
%\textit{RPTU Kaiserslautern-Landau} \\
%Kaiserslautern, Germany \\
%paul.lukowicz@dfki.de}
%}

\author{\IEEEauthorblockN{
        Lala Shakti Swarup Ray\IEEEauthorrefmark{2},
        Bo Zhou\IEEEauthorrefmark{1}\IEEEauthorrefmark{2},
        Sungho Suh\IEEEauthorrefmark{1}\IEEEauthorrefmark{2}\thanks{Corresponding author: sungho.suh@dfki.de},
        Paul Lukowicz\IEEEauthorrefmark{1}\IEEEauthorrefmark{2}
        }
    
    \IEEEauthorblockA{\IEEEauthorrefmark{1}Department of Computer Science, RPTU Kaiserslautern-Landau, Kaiserslautern, Germany}
    \IEEEauthorblockA{\IEEEauthorrefmark{2}German Research Center for Artificial Intelligence (DFKI), Kaiserslautern, Germany}
}

\maketitle

\begin{abstract}
Understanding human-to-human interactions, especially in contexts like public security surveillance, is critical for monitoring and maintaining safety. 
Traditional activity recognition systems are limited by fixed vocabularies, predefined labels, and rigid interaction categories that often rely on choreographed videos and overlook concurrent interactive groups. 
These limitations make such systems less adaptable to real-world scenarios, where interactions are diverse and unpredictable. 
In this paper, we propose an open vocabulary human-to-human interaction recognition (OV-HHIR) framework that leverages large language models to generate open-ended textual descriptions of both seen and unseen human interactions in open-world settings without being confined to a fixed vocabulary. 
Additionally, we create a comprehensive, large-scale human-to-human interaction dataset by standardizing and combining existing public human interaction datasets into a unified benchmark. 
Extensive experiments demonstrate that our method outperforms traditional fixed-vocabulary classification systems and existing cross-modal language models for video understanding, setting the stage for more intelligent and adaptable visual understanding systems in surveillance and beyond.
\end{abstract}


\section{Introduction}
% Talk about how LLM is revolutionizing other fields
Large Language Models (LLM) \cite{touvron2023llama,team2024gemma} have evolved significantly beyond their original role in text processing.
These models have demonstrated their ability to tackle complex tasks that transcend natural language processing, providing invaluable insights into different domains like computer vision, robotics, and software development with tasks including image understanding \cite{li2023blip}, robot control \cite{honerkamp2024language,huang2023instruct2act}, code generation \cite{roziere2023code,nijkamp2022codegen}, sensor data generation \cite{ray2024text} and multi-modal comprehension \cite{zhang2023video,hu2024bliva}. 

% Talk about human interaction and how it is different from HAR
Unlike traditional human activity recognition (HAR) methods \cite{chi2022infogcn,fritsch2024mujo,shi2020decoupled,ray2024har}, human-to-human interaction recognition presents unique challenges for machine learning models, primarily because it requires interpreting complex physical activities within a relational and social context. 
Whereas traditional HAR typically focuses on recognizing individual, repetitive actions, such as walking or running., human-to-human interaction recognition necessitates understanding how two or more individuals coordinate their movements, gestures, and spatial relationships. 
This added layer of complexity means that models must be able to interpret a broader range of scenarios, including variations in personal space norms, cultural differences in gestures, and the specific nature of the interaction itself.
The variability inherent in human interactions makes it difficult for models to generalize across different settings and populations. Traditional activity recognition systems focus primarily on identifying and classifying isolated actions performed by individuals, relying on fixed vocabularies, predefined labels, and rigid interaction categories. 
This narrow focus limits their ability to capture the complexity and variety of human-to-human interactions, making it difficult for these systems to generalize to new or unexpected interactions.
Current methods for recognizing human-to-human interactions face significant challenges, particularly in handling their diversity and complexity \cite{hhi24, yin2023two}. 
Traditional models often need help to adapt to new, unseen scenarios, leading to poor performance in open-world settings where unseen and complex interactions frequently occur. 
These limitations highlight the need for more flexible and adaptive models to better generalize beyond their training data.

% talk about how we use our model and llm as a backbone to solve these gaps
To address these challenges, we propose an open-vocabulary approach using LLMs as the backbone. Unlike traditional systems, our framework leverages the flexibility of LLMs to recognize and describe a broader range of interactions, accommodating new and complex scenarios that fall outside predefined categories \cite{ji2024hargptllmszeroshothuman, xia2023unsupervised}. 
By integrating foundational models such as the Segment Anything Model (SAM) \cite{kirillov2023segment}, Vision Transformer (ViT) \cite{dosovitskiy2021imageworth16x16words}, and LLaMA 2 \cite{touvron2023llama2}, our system can generate detailed and context-aware descriptions of human interactions. 
This allows us to decouple the tasks of segmentation, feature extraction, and interaction description. 
Our approach isolates relevant entities and interactions within images or videos, captures contextual and relational visual features, and generates open-vocabulary textual descriptions of interactions. By aligning visual and textual embeddings within a shared space, our method enables the recognition of seen and unseen human interactions in open-world scenarios.

%results in brief
Our approach has shown promising results across various data sets with a cosine similarity score of 0.63. 
We also developed a custom-scripted classifier built on the LLM-Instruct backbone to further refine and evaluate our model. This classifier generates activity classes that outperform traditional models in classification tasks and is capable of recognizing unseen interactions, which contemporary classification models struggle to achieve, as shown in our evaluation.
Although computationally intensive, the proposed method efficiently generates novel activity classes not present in the training data, making it particularly suitable for complex and dynamic HAR scenarios.

\begin{figure*}[!t]
    \centering
    \includegraphics[width=0.7\linewidth]{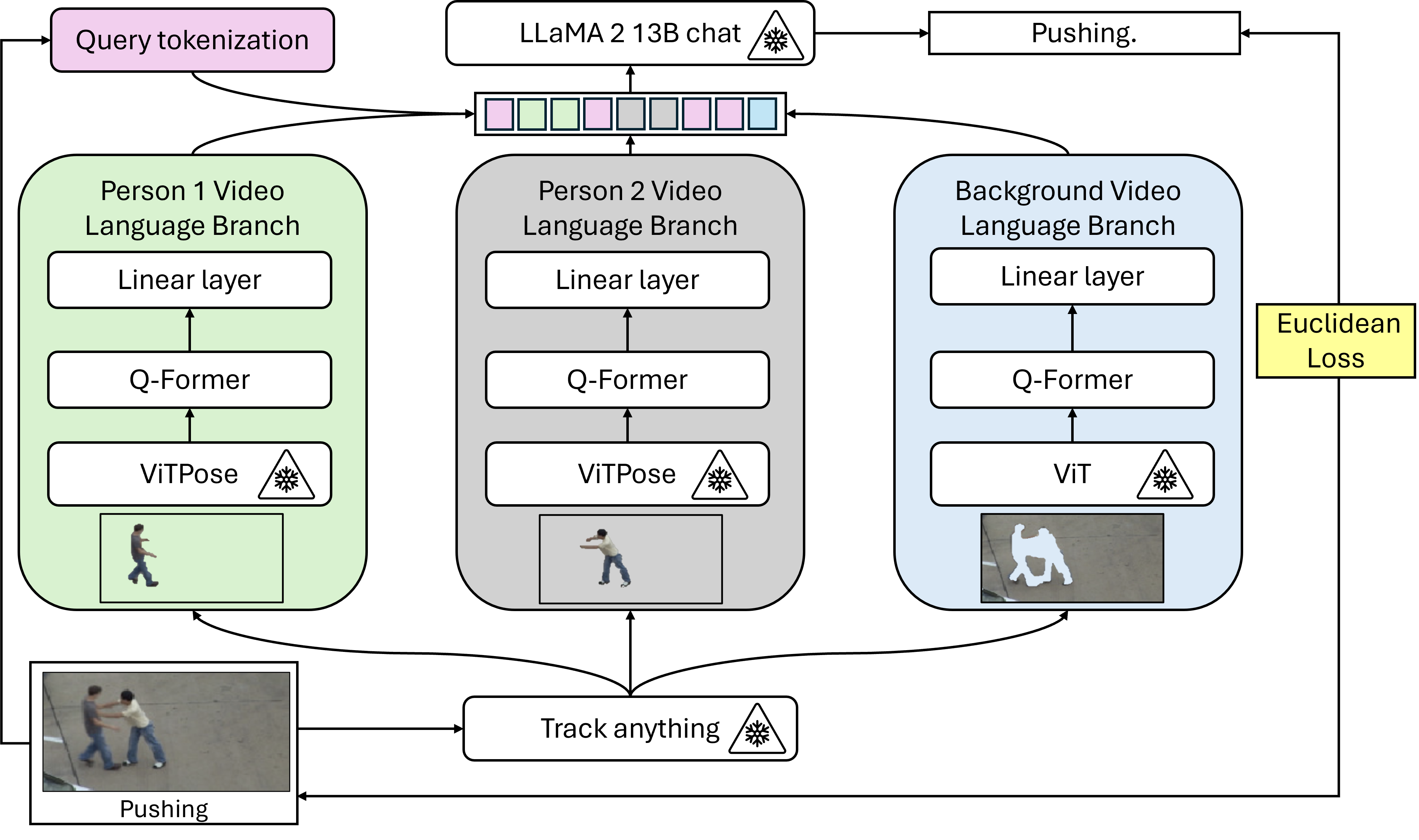}
    \caption{Overview of the proposed OV-HHIR framework that uses three video language branches and LLaMA 2 13B chat to generate open-ended natural language descriptions of human-to-human interactions from dynamic video sequences. During training, Track anything, ViTPose, ViT and LLaMA 2 13B Chat have frozen weights, while the different instances of Q-Former and Linear layers have learnable weights.}
    \label{fig:framework}
    \vspace{-5mm}
\end{figure*}

The key contributions of our research include:

\begin{itemize}

    \item We propose an open-vocabulary human-to-human interaction recognition (OV-HHIR) visualized in \cref{fig:framework} that leverages LLMs to generate open-ended text of human interactions. This approach allows the model to recognize and classify a broader range of activities beyond those explicitly present in the training data.
    % \item Open-Vocabulary Human to Human Interaction Recognition (OV-HHIR): We developed OV-HHIR framework that leverages LLMs to generate open-ended textual descriptions of human interactions. This approach allows the model to recognize and classify a broader range of activities beyond those explicitly present in the training data.

    \item We introduce a comprehensive large-scale unified HHIR chat (HHIRChat) dataset, standardizing human interaction activities from existing benchmark datasets into 103 unique human interaction categories.
    % \item Large-Scale Unified HHIR Chat Dataset(HHIRChat): We introduced and utilized a comprehensive, large-scale dataset specifically designed for HHIR, encompassing a wide variety of human interactions unified to a standard format. This dataset provides a robust foundation for training and evaluating advanced models.

    \item We evaluate our framework both qualitatively and quantitatively by assessing the generated captions and building a classification model to compare its performance with contemporary activity recognition systems. Additionally, we test its open-ended capabilities by identifying interactions it wasn't exposed to during training, a feature lacking in existing models.
    % \item LLM-based Interaction Classification Model: We implemented an HHIR classification model based on our OV-HHIR framework, demonstrating enhanced performance in recognizing complex and dynamic human activities. This model shows significant potential for real-world applications across various domains.

\end{itemize}

\section{Proposed Method}
\subsection{HHIRChat Dataset}
\begin{figure}[!t]
    \centering
    \includegraphics[width=0.9\linewidth]{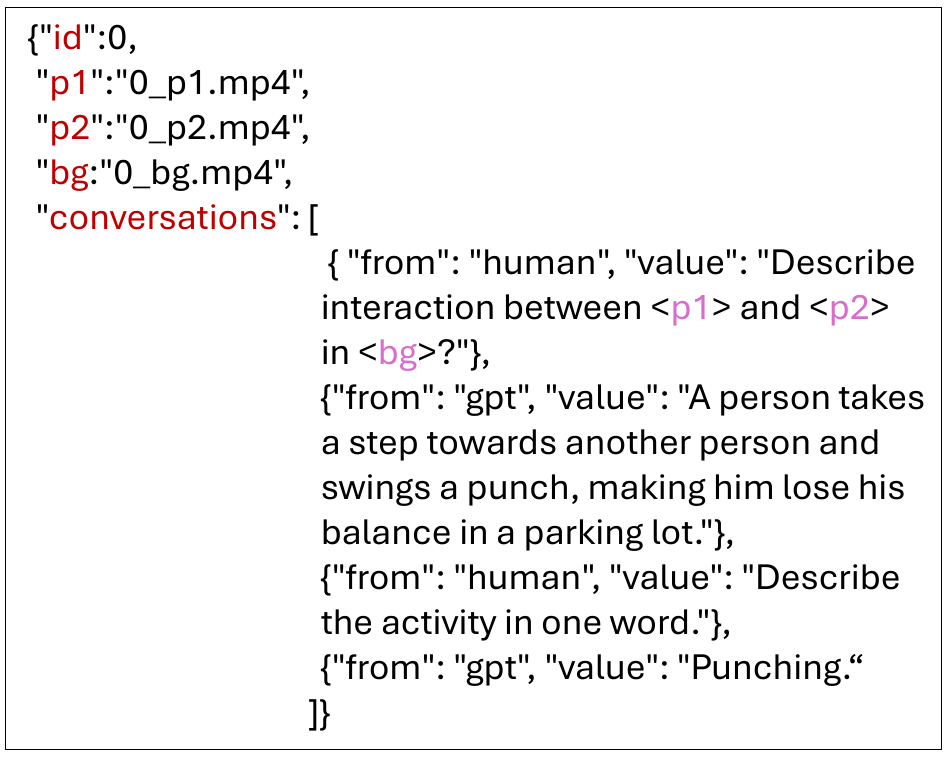}
    \caption{HHIRChat data format that mixes cross-modal tokens.}
    \label{lst:example_interaction}
    \vspace{-5mm}
\end{figure}

To train our model for Human-to-Human Interaction Recognition (HHIR), we aggregated all existing multi-person interaction datasets. Using GPT-4 \cite{achiam2023gpt} and following guidelines similar to those in Text-to-Pressure (T2P) \cite{ray2024text}, we converted hard labels into soft and descriptive labels. For instance, instead of using a simple label \textit{"punching"}, we describe the interaction as \textit{"A person takes a step towards another person and swings a punch, causing them to lose their balance."}
To utilize this data for training, we employed the Track Anything \cite{yang2023track} to create masks for Person 1, Person 2, and the background. 
Subsequently, we applied a simple video masking technique using OpenCV to generate three separate instances of each video: one containing only Person 1, one with only Person 2, and one showing the background scene without either person. 
The processed data is organized in a chat-based format inspired by the VideoChat dataset, as shown in \cref{lst:example_interaction}. 

Our model includes videos sourced from several datasets, such as the Hollywood2 \cite{marszalek2009actions}, SportsHHI \cite{wu2024sportshhi}, TV Human Interaction \cite{patron2010high}, SBU Kinect Interaction \cite{yun2012two}, JPL-Interaction \cite{ryoo2013first}, UT-Interaction dataset \cite{perez2021interaction}, DeepMind Kinetics \cite{kay2017kinetics}, AIR-Act2Act dataset \cite{ko2021air}, ShakeFive2 \cite{eccv-hbu-2014} and the Mutual Actions dataset from NTU RGB+D \cite{liu2019ntu}. 
These datasets cover 103 unique human interaction activities, all of which were converted into soft labels along with 86,623 sequences of videos with dynamic lengths as provided in \cref{tab:data}.

\begin{table}
\caption{Data statistics}
\label{tab:data}
\centering
\footnotesize
\begin{tabular}{c|c|c|c}
Dataset & subjects & Actions & Samples \\
\hline
Hollywood2 & - & 12 & 2,517 \\
UT-Interaction & 6 & 6 & 120 \\
TV Human Interaction & - & 4 & 300 \\
SBU Kinect Interaction & 7 & 8 & 300 \\
JPL-Interaction & 8 & 8 & 312 \\
ShakeFive2 & 33 & 9 & 180 \\
DeepMind Kinetics & - & 8 & 53 \\
NTU RGB+D 120 & 106 & 11 & 6,378 \\
AIR-Act2Act & 100 & 26 & 8,276 \\
SportsHHI & - & 34 & 55,631 \\
\hline
HHIRChat (Ours) & - & \textbf{103} & \textbf{86,623} \\
\end{tabular}
\end{table}

%32+4+6+4+11+10+28+26
\subsection{OV-HHIR Architecture}
Our approach builds upon the foundations of the Video-LLaMA \cite{zhang2023video} and Caption Anything \cite{wang2023caption} architectures, introducing essential modifications that enhance the system’s ability to recognize and describe human interactions within dynamic video streams.
A central innovation is the introduction of distinct vision-language branches for each individual in the video, with an optional branch for background context. This structure enables a more detailed and context-aware analysis of human interactions. Before passing the video to the vision-language branches, we employ the Track Anything model to segment the video into separate streams. This segmentation ensures that each branch focuses on a specific interaction component, allowing for precise and targeted analysis.

We propose three vision-language branches: one for Person 1, one for Person 2, and an optional branch for the background scene. The Person 1 and Person 2 branches process the video segments corresponding to each individual, utilizing pre-trained ViTPose \cite{xu2022vitpose} encoders. Since ViTPose is trained explicitly for human pose estimation in the video, it captures essential features from the human-centered segments more effectively than the vanilla ViT \cite{yuan2021tokens} used in the original Video-LaMA model. After feature extraction, temporal information is introduced via a positional embedding layer, and a Video Q-former aggregates frame-level features into a cohesive representation. Fully connected (FC) layers then resize the output from the Q-former \cite{li2023blip} to match the size of the text embedding.
The optional Background branch processes the environmental context, excluding the individuals. This branch utilizes a pre-trained vanilla ViT, which is more suited for generalized feature extraction and follows the same process as the person branches to generate the final embedding.

The outputs from all branches: Person 1, Person 2, and the background are integrated by projecting their video embedding into a shared space where they align with the text embedding. Combined with other generated embedding, this alignment forms the final input for the LLaMA 2 13B chat \cite{touvron2023llama} model. By aligning video and text embedding in this manner, the modified Video-LLaMA architecture generates open-vocabulary textual descriptions that accurately capture interactions between individuals and their surrounding context. This significantly enhances the model’s ability to recognize and describe complex human interactions.
Our architecture is trained in a multi-branch cross-modal framework, ensuring that each branch learns to represent its respective video segment accurately and contributes to the overall understanding of the interaction. By isolating and analyzing distinct aspects of the video, this architecture provides a robust solution for recognizing and describing intricate human interactions in diverse and dynamic environments.

\section{Experimental Results}

\begin{figure*}[!t]
    \centering
    \includegraphics[width=0.77\linewidth]{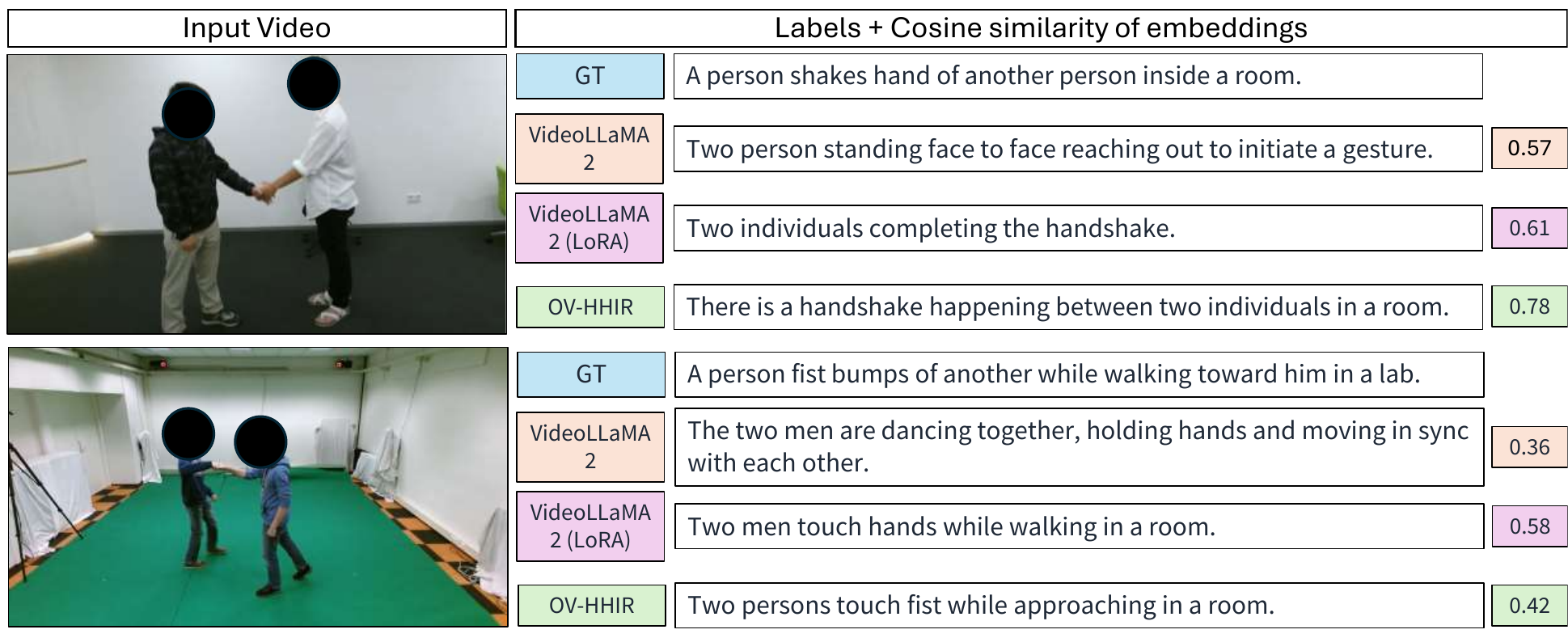}
    \caption{Examples showing the ground-truth vs predictions by Vanilla VideoLLaMA 2, VideoLLaMA 2 LoRA, and OV-HHIR.}
    \label{fig:res}
    \vspace{-5mm}
\end{figure*}

The OV-HHIR model is implemented in PyTorch and trained on a Linux-based cluster with multiple A6000 GPUs.
During the training, all instances of ViTPose, ViT, and LLaMA 2 13B chat are frozen while the remaining instances of Q-Former and the FC layers are trainable. 
The model takes 16 sampled frames from the whole video as input which ensures the system works for dynamic video lengths regardless to total frames as well as the computational need for the model is managable. Cross Entropy loss is calculated between embeddings generated by LLaMA 2 and the ground truth.
% over the final embedding generated by the LLaMA 2 vs the ground truth. 

\subsection{Quantitative Evaluation}
We evaluated our model using cosine similarity, comparing its performance to state-of-the-art (SOTA) cross-modal video-based LLMs like VideoLLaMA and VideoLLaMA 2 \cite{cheng2024videollama}, and LoRA \cite{hu2021lora} adaptations of those models for interaction recognition on our HHIRChat dataset.
As shown in \cref{tab:eva1}, our model significantly outperforms existing SOTA models, achieving a higher cosine similarity score.
% Our model is better than all existing SOTA video-based cross-modal LLMs as provided in \cref{tab:eva1}.
An apparent advantage of our approach is its ability to accurately distinguish interactions between individuals in crowded environments, a task where existing LLMs struggle because they treat the whole frame as a single input.
% Besides the apparent advantage of cosine similarity, our model also recognizes interactions between two persons in a crowd. It is not easy in the existing LLMs, as they take the whole frame as a single input.
To compare our model with contemporary classification models, we used prompt engineering to generate a single class from the limited \textit{N} classes and thereby generate the macro-F1 score for OV-HHIR as given in \cref{tab:eva2}.
while Baseline video classifier \cite{perez2021interaction} and ASL-HAR classifier (uses pre-trained ViT to generate video embedding before classification) \cite{ray2024har} were trained from scratch for each dataset, the OV-HHIR takes the same pre-trained model across the entire HHIRChat dataset.
% We want to address that while the Baseline video classifier and ASL-HAR classifier (uses pre-trained ViT to generate video embedding before classification) \cite{ray2024har} are trained from scratch for each dataset, the OV-HHIR takes the same pre-trained model over HHIRChat dataset.
We observed that while the improvement for OV-HHIR based classifier is minimal for some datasets like AIR-Act2Act, NTU RGB+D, UT-Interaction, and ShakeFive2 where the background setting, viewing angle as well as the number of participants is very limited in cases of datasets like Hollywood 2, TV Human Interactions and DeepMind Kinetics, where the data was collected from YouTube or other video sources with diverse participants, viewing angles, and backgrounds, our model outperforms the contemporary classification models by a large margin.

To evaluate further whether HHIR can truly recognize open-ended human-to-human interaction, we collected a set of 6 uncommon activities, i.e., bowing down, ear whispering, forehead kissing, lifting, and taunting, originally not present in the HHIR chat datasets, from YouTube and used an OV-HHIR-based classifier to classify them into one of the existing 103 interaction classes from the original training set and the six new interactions from this evaluation set.
The model uses a Macro-F1 score of 0.315$\pm$0.023, which is still quite impressive. The same approach would always result in 0 for any classification model since it can not recognize any activity beyond its training set.

\begin{table}
\caption{Comparison of different Video-LLMs with OV-HHIR for the generated embedding using a cosine similarity metric.}
\label{tab:eva1}
\centering
\footnotesize % or use \tiny for smaller text
\begin{tabular}{c|c}
Model & Cosine Similarity (Mean ± Std)\\
\hline
VideoLLaMA  & 0.234 ± 0.051 \\
VideoLLaMA (LoRA HHIRChat) & 0.378 ± 0.042 \\
VideoLLaMA-2 & 0.212 ± 0.056 \\
VideoLLaMA-2 (LoRA HHIRChat) & 0.414 ± 0.033 \\
\hline
OV-HHIR (Ours) & \textbf{0.628 ± 0.019}\\
\end{tabular}
\vspace{-5mm}
\end{table}

\begin{table}
\caption{Comparison of different human-to-human interaction recognition classifiers that includes baseline classifier vs ASL-HAR Classifier and LLM based OV-HHIR classifier.}
\label{tab:eva2}
\centering
\footnotesize
\begin{tabular}{c|c|c}
Dataset & Model & Macro-F1 \\
\hline
& Baseline classifier & 0.671 $\pm$ 0.019 \\
AIR-Act2Act & ASL-HAR classifier & \textbf{0.693 $\pm$ 0.029} \\
& OV-HHIR classifier & 0.688 $\pm$ 0.014 \\
\hline
& Baseline classifier & 0.714 $\pm$ 0.012 \\
NTU RGB+D & ASL-HAR classifier & \textbf{0.738 $\pm$ 0.023} \\
& OV-HHIR classifier & 0.728 $\pm$ 0.051 \\
\hline
& Baseline classifier & 0.825 $\pm$ 0.061 \\
UT-Interaction & ASL-HAR classifier & 0.861 $\pm$ 0.048 \\
& OV-HHIR classifier & \textbf{0.877 $\pm$ 0.036} \\
\hline
& Baseline classifier & 0.588 $\pm$ 0.037 \\
ShakeFive2 & ASL-HAR classifier & 0.612 $\pm$ 0.015 \\
& OV-HHIR classifier & \textbf{0.631 $\pm$ 0.032} \\
\hline
\hline
& Baseline classifier & 0.412 $\pm$ 0.047 \\
Hollywood 2 & ASL-HAR classifier & 0.493 $\pm$ 0.033 \\
& OV-HHIR classifier & \textbf{0.577 $\pm$ 0.042} \\
\hline
& Baseline classifier & 0.662 $\pm$ 0.024 \\
TV Human Interaction & ASL-HAR classifier & 0.694 $\pm$ 0.047 \\
& OV-HHIR classifier & \textbf{0.807 $\pm$ 0.039} \\
\hline
& Baseline classifier & 0.454 $\pm$ 0.021 \\
DeepMind Kinetics & ASL-HAR classifier & 0.471 $\pm$ 0.036 \\
& OV-HHIR classifier & \textbf{0.592 $\pm$ 0.032} \\
\end{tabular}
\vspace{-5mm}
\end{table}

\subsection{Qualitative Evaluation}
In \cref{fig:res}, we used two examples to showcase our results compared to VideoLLaMA 2. 
Purely relying on cosine similarity can lead to incorrect evaluations, as demonstrated in the second example where, despite the ground truth and predictions indicating the same interaction activity, the sentence semantics reduce the cosine similarity compared to VideoLLaMA 2 LoRA, making it inaccurate, which is why we employed a classifier approach for an unbiased evaluation.
\subsection{Limitations}
While open-ended generation models are powerful, they come with challenges such as potential errors and high memory overhead. When the model encounters chaotic video scenes with too many things happening in the scene or constant changes, it may generate irrelevant or nonsensical text, which dilutes meaningful content and leads to misclassification in our LLM prompt-engineering classifier. Since the OV-HHIR relies on several pre-trained foundation models, their limitations also affect it. Despite offering greater accuracy and flexibility, it demands significantly more memory than typical classification models. Further optimization is needed to enable real-time deployment on edge devices.

\section{Conclusion}
In this paper, we introduced OV-HHIR, a novel open-vocabulary human-to-human interaction recognition framework that uses LLM as a backbone to overcome the limitations of traditional activity recognition systems.
While our extensive experiments demonstrated its superior performance over existing methods, challenges such as high memory overhead and occasional errors highlight areas for future improvement, especially for real-time deployment.

In future work, we will extend the framework to support interaction detection across different scenarios, including single-person activities, two-person, and group-based interactions, enabling the system to recognize interactions in a wide range of settings.

\vfill\pagebreak

\bibliographystyle{IEEEbib}
\bibliography{refs}

\end{document}